\documentclass[journal]{IEEEtran}
\usepackage{booktabs}
\usepackage{multirow}
\usepackage{graphicx}
\usepackage{caption}
\ifCLASSINFOpdf
\else
\fi

\usepackage{cite}
\usepackage{authblk}

\begin{document}
\title{Visual-Textual Association with Hardest and Semi-Hard Negative Pairs Mining for Person Search}
\author[1]{Jing Ge}
\author[1]{Guangyu Gao}
\author[1]{Zhen Liu}
\affil[1]{Beijing Institute of Technology}

% \markboth{Journal of \LaTeX\ Class Files,~Vol.~14, No.~8, August~2015}%
% {Shell \MakeLowercase{\textit{et al.}}: Bare Demo of IEEEtran.cls for IEEE Journals}
\maketitle

\begin{abstract}
Searching persons in large-scale image databases with the query of natural language description is a more practical important applications in video surveillance. Intuitively, for person search, the core issue should be visual-textual association, which is still an extremely challenging task, due to the contradiction between the high abstraction of textual description and the intuitive expression of visual images. However, for this task, while positive image-text pairs are always well provided, most existing methods doesn't tackle this problem effectively by mining more reasonable negative pairs. In this paper, we proposed a novel visual-textual association approach with visual and textual attention, and cross-modality hardest and semi-hard negative pair mining. More specifically, the proposed approach mainly featured in three aspects. Firstly, in view of the highly abstract but incomplete description of the text and the comprehensive but non-focused description of the image, both visual and textual attention are involved for more consistency matching. Specifically, we designed a Smoothed Global Maximum Pooling (SGMP) to extract more concentrate visual feature, and also a memory attention based on LSTM's cell unit for more strictly correspondence matching. Secondly, to better distinguish different identity’s both visual and text representation, the network should consider more comprehensive samples, i.e. hard examples. Thus, more valuable negative pairs are mined by defining hardest and semi-hardest pairs in both direct and indirect way. Notably, in the proposed hard negative pairs mining, we focus more on mining cross-modality hard negative pairs for more satisfactory performance. Meanwhile, with such pairs, we combine the triplets loss on single modality and cross entropy loss on the hardest and semi-hard pairs cross modality together. Finally, in order to evaluate the effectiveness and feasibility of the proposed approach,  we conduct extensive experiments on typical person search datasdet: CUHK-PEDES, in which our approach achieves the top1 score of $55.32\%$ as a new state-of-the-art. Besides, we also evaluate the semi-hard pair mining approach in COCO caption dataset, and validate the effectiveness and complementarity of the methods.
\end{abstract}


\IEEEpeerreviewmaketitle

\section{Introduction}

Recently, person-related recognition tasks, e.g., pedestrian detection and person re-identification have attracted much attention with the development of convolutional neural networks(CNNs). When referring to person re-identification, most works target searching for similar individuals from a pedestrian pool, where all the individuals have been already manually cropped out. However, for most practical cases, such a pedestrian pool cannot be well prepared, and we need to search the candidates directly from the whole video frames. Therefore, person search as a more practical task, is defined as the combination of pedestrian detection and re-identification, and can be applied to many important applications (e.g. multi-camera tracking and activity analysis). Nowadays, most cities are usually equipped with thousands of surveillance cameras which generate gigabytes of video data every second. Searching persons in such large-scale databases with a specific query has very practical applications, but also face problems of efficiency and accuracy. Nevertheless, in many practical scenarios, person search with query of text, seems to be more realistic. For example, people can usually describe the characteristics of a person who needs to be searched verbally quickly, but mostly it’s not always available to provide a corresponding picture. 

In addition, although it has wide applications in area such as video surveillance, searching person in a database with free-form natural language description is also a very challenge task. Therefore, some researches proposed methods to deal with such challenges for more practical applications. Li et al. \cite{li2017person} studied the problem of person search with natural language description. Most importantly, they provided the CHUK Person Description Dataset (CUHK-PEDES) by collecting a large-scale person description dataset with detailed natural language annotations and person samples from various sources. Meanwhile, the authors in \cite{Li_2017_ICCV} proposed an identity-aware two-stage framework for the textual-visual matching, in which, the first stage learned identity-aware representation, and it matched salient image regions and latent semantic concepts for textual-visual affinity estimation in stage 2. Chen et al. \cite{Chen_2018_ECCV} exploited natural language description as additional training supervisions for effective visual representation by global and local image-language association.

Actually, person search with query of text, is also very related to visual-textual association, namely, the core problem is to provide better visual-textual embedding and matching\cite{Chen_2018_ECCV} \cite{faghri2017vse++}. However, for visual-textual association, the crucial difficulty should be the contradiction between the high abstraction of the textual description and the intuitive expression of images. Therefore, in visual-textual association based person search, the key points should be: the most suitable visual and textual feature embedding, and the most reasonable matching metric. Occasionally, these two key points are also combined together by methods such as metric learning. For example, Faghri et al. proposed the VSE++\cite{faghri2017vse++}, which introduce a simple change to common loss functions used for multi-modal embedding.

Because of the highly abstract but incomplete description of the text and the comprehensive but non-focused description of the image, we exploit more reasonable visual-textual association with visual and textual attention and cross-modality hard pair mining for person search. Because the core purpose of visual text association based person search or is to measure the similarity between visual and textual features. The inconsistency between the expression of visual features and text features makes traditional visual feature extraction often contain many features that are not described by text features. At this time, it is often problematic to calculate the distance or similarity between text and image feature vectors by various distance measures.

In this paper, the overall architecture is designed based on the CNN networks (i.e. ResNet\cite{he2016deep}) to extract the visual feature, and the bi-LSTM to extract the textual features. Firstly, in order to make the visual features focus more on discriminate characteristics, i.e. saliency part, a Smoothed Global Maximum Pooling (S-GMP) module is designed to extract the visual features more consistent with text description. In person search, visual-textual association concerns more on common features of both modalities for more reasonable similarity measurement. It means that the visual feature should concentrate on more salient local features. While the traditional Global Average Pooing (GAP)\cite{lin2013network} in CNN extracted more general or smoothed features implicitly, we designed the S-GMP with Global Max Pooling (GMP) but weighted by sigmoided GAP. This makes the GMP focus more on local salient features but smoothed by GAP with sigmoid function.

While the textual features are extracted by the bi-LSTM network, the last cell unit always stores the global attentions of the whole sentence. For more consistent cross modality feature matching, we also designed the memory attention based on the cell unit of bi-LSTM to weight the similarity. In details, the output of the last cell will transfer to attention and multiply to the traditional visual-textual association similarity. Therefore, considering such a weighted memory attention, it achieves more strictly correspondence between visual and textual feature vector.

Besides, our goal is to distinguish different person’s sample but cluster sample from the same one. Therefore, to better distinguish different person’s both visual and text representation, we propose a new loss function which combine the triplet loss on single modality and the hard pair mining based cross entropy loss on cross modality. Notably, selecting the hardest negatives pair may make the model fall into bad local minima early in training, and it can result in a collapsed model especially. Therefore, we defined two types of hard negative pairs, namely, the hardest negative sample to the anchor itself, and the semi-hard negative sample that is the hardest sample to the anchor’s teammate. In such a hard negative pair mining, we focus more on mining cross-modality hard samples for more satisfactory performance. Finally, we conduct extensive experiments on CUHK-PEDES, and our approach achieves the top1 score of $55.32\%$, which is a new state-of-the-art.

All in all, the main contributes of our approach could be summarized as:
\begin{enumerate}
\item Both the visual and textual attention, namely, the S-GMP in the CNN and the cell memory attention from the bi-LSTM are involved in for more consistent feature representation and corresponding matching.
\item To better distinguish different person’s both visual and text representation, we combine both the triplet loss on single modality and cross-modality hardest and semi-hard pair mining loss between visual and textual modality.
\item Extensive ablation studies validate the effectiveness and complementarity of the proposed methods, and the results show that our approach achieves state-of-the-art performance in CUHK-PEDES.
\end{enumerate}

\section{Related Work}
Cross-modality retrieval is designed to use samples within a specific modality e.g. text to search for samples of approximate semantics within another modality e.g. image.For Text-Image cross-modality retrieval,some brilliant models have been proposed. Feng et al.\cite{feng2014cross}used the encoder and decoder of autoencoder to learn the correlation between different modality and the representation of the samples respectively.Vendrov et al.\cite{vendrov2015order}employed a hierarchy order for image and text.Using a similar method as Vendrov et al.,Wang et al.\cite{wang2016learning} proposed a model which introduced additional hard triplet loss and Structure-Preserving to reach a new state-of-the-art.Also made a change to common loss functions Faghri et al.\cite{faghri2017vse++} made full use of triplet loss and implement an end-to-end training network.As attention mechanism has shown its advantages,more and more work employ it to help improving the performance of models and latest work by Ji et al.\cite{ji2019saliency}is an example.

Meanwhile,as an independent computer vision task,person search has also attracted more and more attention.Person search has some relevance to another two vision task pedestrian detection and person re-identification(re-id).Pedestrian detection solves the problem of finding pedestrians in an outdoor scene.Early person detection models,for example,\cite{benenson2012pedestrian}\cite{benenson2013seeking}\cite{hariharan2012discriminative} used artificially designed features such as ICF\cite{dollar2009integral},HOG\cite{dalal2005histograms},Haar\cite{papageorgiou1998general}features for pedestrian search.With the development of convolutional neural networks(CNNs),more and more methods based on learned features have been proposed such as \cite{li2017scale}\cite{mao2017can}\cite{Zhou_2018_ECCV}.Different from pedestrian detection,person re-id\cite{Li_2017_CVPR}\cite{Chen_2017_CVPR}\cite{fu2019self}\cite{qi2019novel}places an emphasis on matching the cropped person image between the query and the gallery.However in practical cases,cropped person images cannot be obtained directly.Thus person search has more practical applications and it adopts a method which combines pedestrian detection and person re-id normally.Xiao et al.\cite{xiao2016end} proposed a network which solved pedestrian detection and person search in an end-to-end manner.On this basis,they proposed a method\cite{xiao2017joint}in which a new loss function called OIM loss was introduced to save the network from updating the large weighted classification matrix.Later Xiao et al.\cite{xiao2017ian}considered increasing the similarity between different instances which belong to the same identity and a center loss is employed to realize this idea.Further experiment has proved the validity of this method.Different from the method of combing pedestrian detection and person search,Liu et al.proposed a new model called neural person search machine(NPSM)\cite{liu2017neural}which recursively focuses on a subregion of previous step and finds the target person at last.This work provide a new idea about person search.A recent work done by Yan et al.\cite{yan2019learning}propose to mine the contextual information of the target person to improve the performance of person search and a specific graph structure was designed to model contextual information.

Both Text-Image cross-modality retrieval and person search are current mainstream research directions.Contrastively, person search with natural language descriptions is not popular, it's more concerned with learning the attributes of pedestrians such as appearance, behavior. Text-Image cross-modality retrieval focuses more on the detection and recognition of objects and distinguishes the relationship between objects. Therefore, Early language datasets for vision can't be used to train person search algorithms with language descriptions. Li et al.\cite{li2017person} collected a large-scale language description dataset CUHK-PEDES with rich language annotations and conducted the user studies. At the same time, a recurrent neural network(RNN)\cite{zaremba2014recurrent} with gated neural attention was proposed. The feature vector from the visual network is used in the network to generate text features, to learn the attention of each word to the content of image, and to learn a control gate as the weight of the word.

Li et al.\cite{li2017person} proposed a two-stage framework for Text-Image retrieval. In the first stage, text and visual features are embedded into the CMCE loss, and sample pairs that are easy to distinguish are selected, and an initial point is provided for the second-stage network. In the second stage of the network, a new potential common attention mechanism is proposed, which corresponds to the area of the image. Zheng et al.\cite{zheng2017dual} used convolutional neural networks to extract features for both text and image networks\cite{zheng2017dual}. At this time, adding ranking loss to instance loss to train the entire network.They propose a more granular approach to retrieving text with the most matching image by embedding Text-Image. Chen et al.\cite{Chen_2018_ECCV} used the identity tags to learn the relationship between global images and languages, and used local image regions and noun phrases to construct local relationships.

However, they did not conduct an in-depth discussion on the selection of negative pairs, and design complex algorithms to guide network to learn the relationship between text and images more accurately.But choosing the right negative pairs enables the network to automatically learn which areas of the text and image to focus on, and defeat the artificially designed algorithm even when the dataset is good enough.This paper is devoted to selecting better Text-Image pairs to help the network to train better.

\section{Network Design}
\begin{figure*}[h]%%图
\centering
\includegraphics[width=1\linewidth]{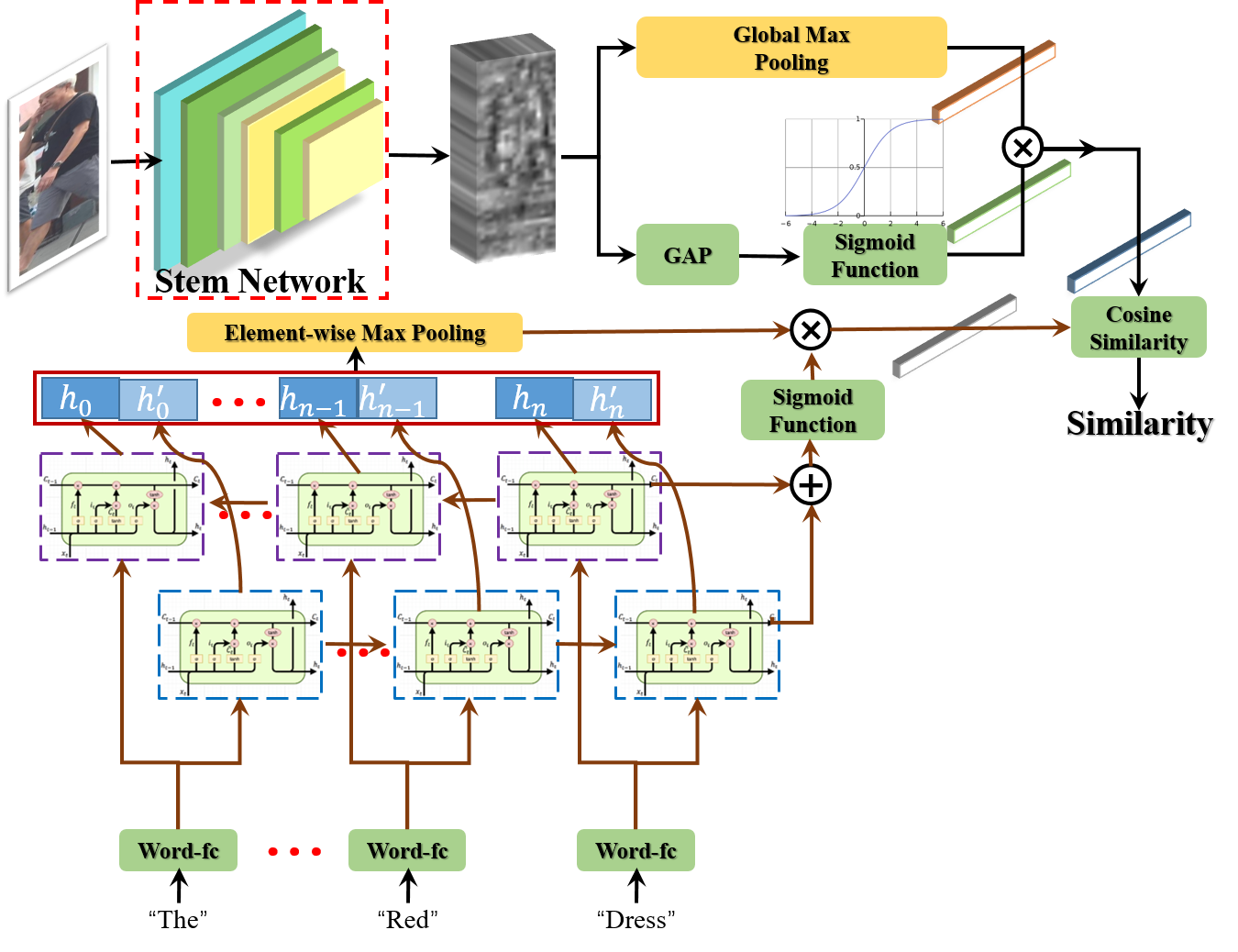}
\caption{The framework of our network.The revised backbone network ResNet-50 extracts visual feature from the input image,then the visual feature is obtained by a Smoothed Global Maximum Pooling (S-GMP) module.Bi-LSTM extracts the features of each word and the cell memory of the entire sentence.All word features generate text features by maxpooling and cell memory attention.Finally, using cosine similarity to measure Image-Text similarity.}
\end{figure*}

Given a dataset \(\mathcal{D}={(I_n, T_n)}^N_{n=1}\), where each tuple contains an image $I_n$ and a textual description $T_n$ for this image, and $N$ is the number of such tuples. The goal of Text-Image association based person search is to retrieve the corresponding person’s image with a query of text. For each pair, the most direct way is to extract both visual and textual features, and use similarity measure to judge the similarity between such visual feature and the corresponding textual feature. As shown in Figure 1, we use the CNN as backbone for visual feature extraction, and bi-LSTM for textual feature extraction. Due to the contradiction between the high abstraction of textual description and the intuitive expression of images, the visual and textual attention (i.e. the S-GMP and the memory attention) are considered for more corresponding features. To learn a more consistent relationship between visual and textual features, we focus on the selection of hard negative samples mining for training. More specifically, we design both the triplets loss on single modality and cross-modality hardest and semi-hard pair mining between visual and textual modality.

\subsection{S-GMP for Visual Feature}
The visual features $\phi(I)$ are obtained by the revised CNN backbone network of ResNet-50. In the original ResNet-50, the global average pooling is used after the last convolutional layer. While the Global Average Pooling (GAP) output the average activation of each feature map, the Global Max Pooling (GMP) will activate on more local saliency, which is very likely to correspond to the highly abstract textual description. Therefore, the GAP is replaced to the GMP firstly. For more smoothed visual feature, the output of the GAP are transfer to range $[0,1]$, just like the probability refer to weight of different feature, and then multiply back to the GMP as visual attention. Actually, with such a Smoothed GMP (S-GMP) in the above manner, it makes the visual features more sharp and focuses more on certain local saliency parts. While the textual description is always sparse or scattered but focus on saliency region, the above-mentioned visual feature will match more to the characteristics of the textual description.

\subsection{Memory Attention for Similarity}
 Each word in the text is represented as a D-dimensional one-hot vector,where $D$ is the vocabulary size.And then the one-hot vector is projected to the word embedding: $e_t=W_e\times O_t$ where $W_e$ is the matrix in which the $ith$ row represents the vector of the $ith$ word in vocabulary.The text is sent to the Long Short-Term Memory network (LSTM) word by word.The LSTM unit embeds the current word into the $e_t$.Using $e_t$ and hidden state $h_t$ of the last step as input,LSTM updates the cell state $c$, and outputs the next hidden state $h_(t+1)$.The hidden state of the last time step is to describe the effective generalization of the text description.After all the words have been processed by LSTM,the hidden state of the output of $N$ words $\{h_n\}_{n=1}^N$ is obtained.The $i_{th}$ dimension of the text feature can be expressed as $\theta(T)_i=\max(h_{ni} )_{n=1}^N$, which makes each dimension of the feature focus more on a particular word.

There is an unequal relationship between text and images. The information contained in the image is very rich, and the text only describes a part of the image. If the image contains information that is not in the description of the text, this will affect the final match. Therefore, we should refine the problem of whether the image and text is matched to whether the feature of the text description exists in the image. In this paper, we use the attention mechanism to pay attention to the features of the text description. We use the cell state of LSTM to focus on the important words in the text.Using sigmoid to map information into $(0,1)$, the possibility that each dimension of the text is described is obtained.

\subsection{Visual Textual Association}
We use the cosine similarity to establish the association between the visual feature $\phi(I)$ and the text feature $\theta(T)$.And then the result is transfer to range $(0,1)$ by sigmoid function to get the final joint representation $s(I,T)=\sigma(c\times\theta(T)\cdot\phi(I))$, where $\sigma$ is the sigmoid function and $c$ is the memory attention. We hope that if $I$ And $T$ belong to the same person $s (I, T)$ is close to $1$, if not the same person $s(I, T)$ is close to $0$. Therefore, we apply a binary cross entropy loss on the score.
\begin{equation}
\mathcal{L}=\frac{1}{N}\sum_{i,j}l_{i,j}\log(s(I_i, T_j))+(1-l_{i,j})\log(1-s(I_i, T_j))
\end{equation}
where N represents the number of Text-Image pairs.And if the image and the text describe the same person, $l_{i,j}=1$, otherwise $l_{i,j}=0$. 

\section{Loss Function Design}
For the problem of cross-modality retrieval,network model in previous work face two problems.First,previous model usually only has a single loss function to measure the matching degree of Text-Image. For a large and complex network, single loss function is difficult to effectively guide the training of the network. Second,they did not fully exploit the information in the negative pairs thus the performance of the model is not satisfactory.In order to solve these two problems,we design a new powerful loss function to guide the learning of our model.

\subsection{Triplet Loss in Single Mode}
Inspired by previous work,we know that increasing the discrimination of the extracted features of the same mode could help the network to learn better.Therefore,to solve the first problem,for each single mode we add a loss function to make the extracted features more distinguishable. For example, the text distance describing the same pedestrian is shorter than the text distance describing different pedestrians. And the same is true for pedestrian images.

We use triplet loss to achieve this, while using Euclidean distance to measure the distance between two feature vectors. Triplet Loss is for each sample $a$ looking for the sample $p$ with the same pedestrian IDs and the sample $n$ with different pedestrian IDs. These three samples make up a triple $(a,p,n)$. For example, $p=\mathop{\arg\min}\limits_{i\in P}E(a,i)$ and $n=\mathop{\arg\min}\limits_{j\in N}E(a,j)$, where $E$ is the Euclidean distance, $P$ is a set of text describing the same person as the text $a$, but does not include $a$, and $N$ is a set of texts that are not the same person as the text $a$.

According to the above methods, for the $i_{th}$ positive sample pair $(I_i,T_i)$, image $I$ and the text $T$, hardest triples are constructed respectively, and the triples $(I_i, I_{ip},I_{in})$ and $(T_i,T_{ip},T_{in})$ are obtained, where $I_{ip}$ is the image with the same person ID as the image $I_i$, $I_{in}$ is the image with the different person ID from the image $I_i$.
The triple of the text is the same.The triplet loss is defined as follows:
\begin{equation}
L_{tri-img}=\frac{1}{N} \sum_i max(\alpha+E(I_i, I_{in})-E(I_i, I_{ip} ),0)
\end{equation}
\begin{equation}
L_{tri-txt}=\frac{1}{N} \sum_i max(\alpha+E(T_i,T_{in})-E(T_i,T_{ip} ),0)
\end{equation}
\begin{equation}
L_{triple} = L_{tri-img}+L_{tri-txt}    
\end{equation}
where $\alpha$ is the margin. The $E$ function is used to calculate the Euclidean distance of two vectors. The triplet loss enables the network to learn the relationship between text and text and between images and images better.

\subsection{Hardest and Semi-Hard Cross-Modality Pairs Mining}
This section explains how we solve the second problem.It is well known that training a cross-modality network requires both positive sample pairs of Text-Image matching and negative sample pairs of Text-Image mismatches. Any text and image describing different pedestrians can make up a negative sample pair but most of which are easy to distinguish.Not only do these pairs not help with the training of the network, they also interfere with the learning direction of the network and weaken the proportion of the loss of excellent samples. Therefore, it is not advisable to randomly select the negative sample pair. 
 
 In order for the network to converge faster and learn more useful information, it is easy to think of selecting indistinguishable negative sample pairs. Given a positive sample pair $(I,T)$, the hardest negatives are given by $I_{ih}=\mathop{\arg\max}\limits_{j}s(I_j,T_i)$ and $T_{ih}=\mathop{\arg\max}\limits_{j}s(I_i,T_j)$,Where $i\neq j$ .hardest negative pairs mining loss function can be obtained as follows:
\begin{equation}
L_{hardest}\!=\!\frac{1}{N} \sum_i [\log(1\!-\!s(I_i,T_{ih}))\!+\!\log(1\!-\!s(I_{ih},T_i ))] 
\end{equation}

The purpose of network learning is to make the matching degree of positive sample pairs larger, and the matching degree of negative sample pairs becomes smaller. However, at the beginning of the training, because the features of the network extraction are not differentiated, if only hardest negative pairs mining is performed, the network cannot specifically pay attention to which dimension of Text-Image vectors should be more different, which may lead to the network falling into local optimization and makes it difficult for the model to converge. As we all know, in hard pairs, the Text-Image is partially matched. if we can make the network pay attention to the different parts of the feature vector, it can make the network converge faster and better.Although the single modality triplet loss function is very helpful for feature extraction,it lacks detailed guidance. For example, the distance between images of the same pedestrian is shorter, but that`s because the network pays attention to the background of the image is the same. Therefore, we also need to use new cross-modality loss function to guide the network to focus on information that is more useful for Text-Image matching. 
    
Given a positive sample pair $(I_i,T_i)$, taking the image $I_i$ as an example, when constructing the triple, we first find a image $I_in$ whose ID is different from $I_i$'s ID,  but whose feature vector is the closest to the image $I_i$'s feature vector, taking $(I_{in},T_i )$ as the negative sample.  Because $I_i$ and $I_{in}$ are similar, the network will pay more attention to the different parts of $I_i$ and $I_{in}$ in the process of learning, and will guide the network to pay more attention to the extraction of different parts when extracting features. Therefore, for each positive sample pair $(I_i,T_i)$ we can generate two negative sample pairs $(I_i,T_{in})$ and $(I_{in},T_i)$, the loss function is defined as follows:
\begin{equation}
L_{semi}\!=\!\frac{1}{N} \sum_i[\log(1\!-\!s(I_i,T_{in} ))\!+\!\log(1\!-\!s(I_{in},T_i ))] 
\end{equation}

\begin{figure}[h]%%图
\centering
\includegraphics[width=1\linewidth]{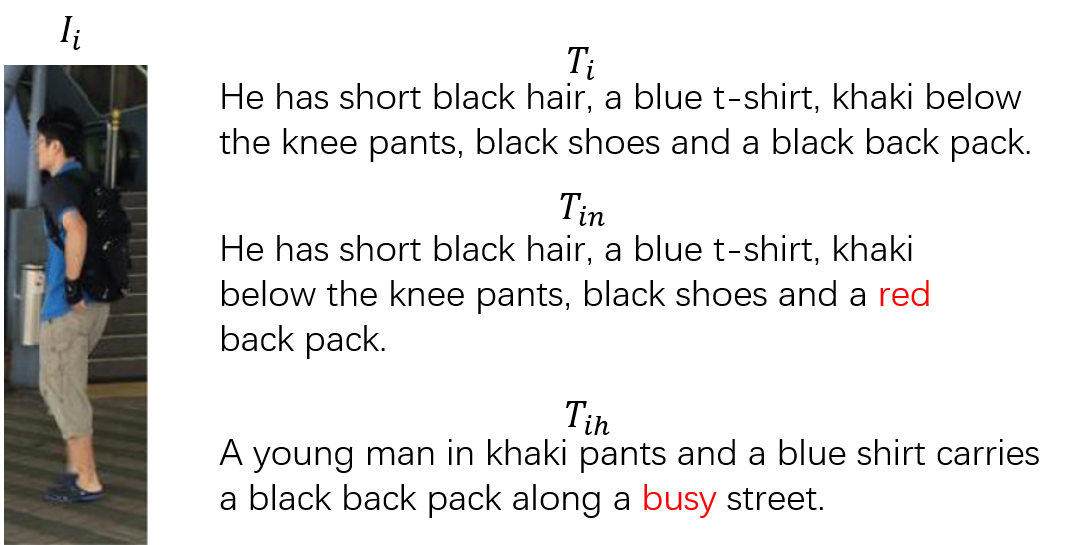}
\caption{Example}
\end{figure}
Current dataset contains limited information. If you only train the network by semi-hard cross-modality pairs mining, you can't make the network learn all the information. As shown in the figure below, for the text $T_i$, according to the rules of semi-hard cross-modality pairs mining, it is easy to select $T_{in}$. Because $T_{in}$ and $T_i$ have only one word "red" different, the network can easily learn the relationship between the word "red" and the image, but with the study of the network, the word "red" has been easily distinguished by the network, so the negative sample pair $(I_i, T_{in})$ is not the best negative sample pair, and the training in the late stage of the network is weak. On the contrary, the hardest negative sample is very helpful for the later training of the network. For this example, when we select the hardest sample, because the word "busy" rarely appears in the dataset, the network will think that the text $T_{ih}$ matches the image $I_i$ more than the text $T_{in}$. The pair $(I_i,T_{ih})$ makes the network pay attention to the relationship between the word "busy" and the image. Therefore, the hardest pair of samples is selected, which still plays an important role in the current lack of datasets. We have added in the selection of negative samples, which makes the network training more accurate and not fall into the local best. The final loss function is:
\begin{equation}
L_{pos}\!=\!\frac{1}{N} \sum_i[\log(s(I_i,T_i ))] 
\end{equation}
\begin{equation}
L=L_{triplet}+ L_{hardest}+L_{semi}+L_{pos}    
\end{equation}

\begin{figure}[h]%%图
\centering
\includegraphics[width=0.5\linewidth]{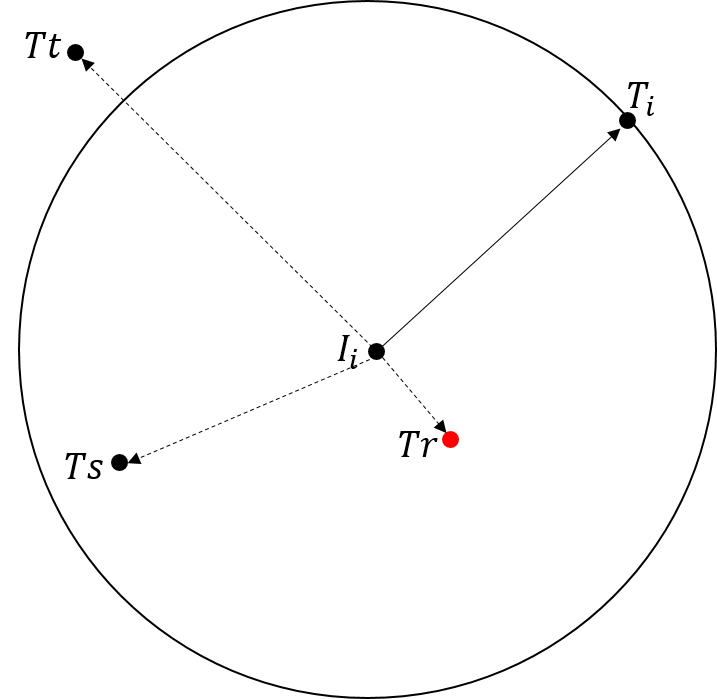}
\caption{Hardest pair mining.$(I_i,T_i)$is a positive pair,other text sample $T_r$,$T_s$,$T_t$ are text which don`t match to $I_i$.And in the figure,the closer the dots representing the image are to the dots representing the text, the more similar the text is to the image.$I_i$ and $T_r$ don`t match,at the same time they have the highest similarity,so they make up a hardest negative pair.}
\end{figure}

\begin{figure}[h]%%图
\centering
\includegraphics[width=0.6\linewidth]{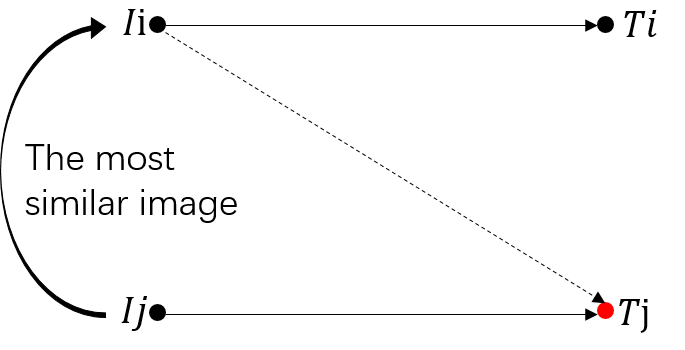}
\caption{Semi-hard pair mining.Both $(I_i,T_i)$ and $(I_j,T_j)$ are positive pairs.$I_j$ is the most similar image to $I_i$,thus$(I_i,T_j)$ make up a semi-hard negative pair.}
\end{figure}

\section{Experimental Results}
\subsection{Implementation Details}
The network training uses the Adam optimizer, the initial learning rate is $2\times 10^{-3}$, decay in the 70th and 90th periods, the decay rate is 0.1.We organize the training batch as follows.Each batch contains a sample of 64 randomly selected people, each person selects two images and each image selects two text descriptions, so each batch contains 128 images of 64 people's and 256 text descriptions. During the training, because there are overlaps between the two negative sample selection methods, we will carry out the semi-hard cross-modality pairs mining firstly.When we select the hardest negative pair, if the hardest pair has been selected as a semi-hard cross-modality pairs, we will select the second hardest negative sample pair to increase the diversity of each batch sample. In the test, the visual feature vectors of all the images are calculated, and the cosine similarity of all the images is calculated for each text, and sorted by the cosine similarity.

The dataset uses the pedestrian re-identification dataset CUHK-PEDES with a natural language description, containing 40,206 images of 13,003 people from the existing pedestrian re-identification dataset. Two independent staff members on the Amazon Labor Crowd-sourcing Platform (AMT) described each person's image in two sentences. In terms of vision, the image of the person from the various re-identification datasets is in different scenes, perspectives and camera specifications, which increases the diversity of image content. In terms of language, the dataset has 80,412 sentence descriptions, including rich vocabulary, phrases, sentence patterns and structures. The labeler has no restrictions on the language describing the image.

\subsection{Comparison with state-of-the-art methods}
 As we can see from the table \uppercase\expandafter{\romannumeral1}, our method that using VGG16\cite{simonyan2014very} and ResNet-50 as the backbone of visual network both produce state-of-the-art performance on CUHK-PEDES. At the same time, we also carried out a wide range of ablation studies to verify the effectiveness and complementarity of the proposed method. We focused on S-GMP in CNN and cell memory attention from bi-LSTM. In addition, we also studied the triplet loss\cite{schroff2015facenet} on single modality and cross-modality hardest and semi-hard pairs mining between visual and textual modality.

\begin{table}[htbp]
  \caption{Performance on COCO Caption}
  \centering
  \setlength{\tabcolsep}{5.3mm}{
    \begin{tabular}{cccc}
    \toprule
    \multirow{2}[4]{*}{Methods} & \multicolumn{3}{c}{CUHK-PEDES} \\
\cmidrule{2-4}          & top-1 & top-5 & top10 \\
    \midrule
    GNA-RNN (VGG-16) & 19.05 & --    & 53.64 \\
    IATV(VGG-16) & 25.94 & --    & 60.49 \\
    DPCE(ResNet-50) & 44.4  & 66.26 & 75.07 \\
    GDA+LRA(ResNet-50) & 43.58 & 66.93 & 76.26 \\
    Ours(VGG-16) & 50.57 & 74.12 & 81.82 \\
    Ours(ResNet-50) & \textbf{55.32} & \textbf{77} & \textbf{84.26} \\
    \bottomrule
    \end{tabular}}%
  \label{tab:addlabel}%
\end{table}%

\subsection{Ablation Experiments}

\begin{table}[htbp]
  \centering
  \caption{Hard and Semi-hard Pairs Mining}
  \setlength{\tabcolsep}{5.3mm}{
    \begin{tabular}{cccc}
    \toprule
    \multirow{2}[4]{*}{Methods} & \multicolumn{3}{c}{CUHK-PEDES} \\
\cmidrule{2-4}          & top-1 & top-5 & top10 \\
    \midrule
    Semi-hard & 50.7  & 73.49 & 81.86 \\
    Triplet+Semi-hard & 53.01 & 75.07 & 83.3 \\
    Triplet+Semi-hard+hard& \textbf{55.32} & \textbf{77} & \textbf{84.26} \\
    \bottomrule
    \end{tabular}}%
  \label{tab:addlabel}%
\end{table}%

\begin{table}[htbp]
  \centering
  \setlength{\tabcolsep}{7.3mm}{
  \caption{GMP for Visual Feature}
    \begin{tabular}{cccc}
    \toprule
    \multirow{2}[4]{*}{Methods} & \multicolumn{3}{c}{CUHK-PEDES} \\
\cmidrule{2-4}          & top-1 & top-5 & top10 \\
    \midrule
    Avgpool   & 54.12 & 75.7  & 83.91 \\
    Maxpool   & 54.38 & 76.27 & 84.24 \\
    S-GMP & \textbf{55.32} & \textbf{77} & \textbf{84.26} \\
    \bottomrule
    \end{tabular}}%
  \label{tab:addlabel}%
\end{table}%

\begin{table}[htbp]
  \centering
  \setlength{\tabcolsep}{6mm}{
  \caption{Memory Attention for Textual Feature}
    \begin{tabular}{cccc}
    \toprule
    \multirow{2}[4]{*}{Methods} & \multicolumn{3}{c}{CUHK-PEDES} \\
\cmidrule{2-4}          & top-1 & top-5 & top10 \\
    \midrule
    Final Hidden  & 50.88 & 74.19 & 81.64 \\
    Avgpool   & 49.45 & 72.15 & 80.77 \\
    Maxpool   & 53.89 & 75.8  & 83.7 \\
    Memory Attention & \textbf{55.32} & \textbf{77} & \textbf{84.26} \\
    \bottomrule
    \end{tabular}}%
  \label{tab:addlabel}%
\end{table}%

\begin{table}[htbp]
  \centering
  \setlength{\tabcolsep}{5mm}{
  \caption{Dropout}
    \begin{tabular}{cccc}
    \toprule
    \multirow{2}[4]{*}{Methods} & \multicolumn{3}{c}{CUHK-PEDES} \\
\cmidrule{2-4}          & top-1 & top-5 & top10 \\
    \midrule
    Without dropout  & 54.00 & 76.02 & 84.01 \\
    Both dropout   & 53.87.45 & 76.59 & 84.11 \\
    Dropout for positive & \textbf{55.32} & \textbf{77} & \textbf{84.26} \\
    \bottomrule
    \end{tabular}}%
  \label{tab:addlabel}%
\end{table}%

Table \uppercase\expandafter{\romannumeral2} shows that the triplet loss is very helpful for the training of the network. Its effect is similar to SVM (Support Vector Machine), which makes the features extracted from text and images more robust. Because the selected triples are the most difficult triples, the model can better learn synonyms, as well as synonymous statements of different expressions when shortening the distance between similar texts. What's more, expanding the distance between texts describing different people, more attention can be paid to words and phrases with different semantics, which makes the learning of the network more targeted. Negative sample selection in cross-modality hardest and semi-hard pair mining depends on single-modality, and better cross-modality model will improve the triple selection in single-modality.Therefore, the above two methods promote each other.

Maxpooling in visual network can better use the relationship between the image area and the text.However, for a single sample, only a few features of high-dimensional image features are effective, so maxpooling will introduce a lot of noise.Table \uppercase\expandafter{\romannumeral3} shows the effectiveness of Smoothed Global Maximum Pooling.

Similarly,Maxpooling in text network can better use the relationship between the word and the image area.Text features can be obtained by weighting word features or the last hidden layer feature of bi-LSTM.Table
\uppercase\expandafter{\romannumeral4} shows that Compared with the last hidden layer features and avgpooling features, maxpooling features have a significant effect.A text feature contains less information than a visual feature, so there is a more serious noise problem for taking the maximum for each dimension.therefore,the cell memory attention from the bi-LSTM shows good results in experiments.

Dropout \cite{srivastava2014dropout} is a method for optimizing artificial neural networks with deep structure. In the learning process,dropout is realized by reducing the partial weight or output of the hidden layer to zero and reducing the inter-node co-dependence. Regularization reduces its structural risk. Because the selected negative sample pair is a difficult sample, the text in the negative sample pair is partly matched to the image part. When performing a dropout operation, it is very likely that the unique neurons representing the differences of Text-Image will be set to 0, Causing text and visual features to match each other after dropout. In this way, the network cannot learn valid information in the negative sample pair, and even guides the network to learn in the wrong direction. Therefore, only the feature vector in the positive sample pairs are processed using dropout, and the dropout operation is not performed in the negative sample pairs.

\section{Conclusions}
In this paper,We have studied and improved the traditional natural language description search pedestrian network framework, including text and visual network feature extraction, cross-modal sample pair mining and loss function research. The independent extraction features of the two networks make the network highly efficient. At the same time, the sample pair mining technology is suitable for all cross-modal retrieval problems, and has a wide application prospect.
\bibliography{ref}
\bibliographystyle{IEEEtran}

\end{document}